\documentclass[10pt,twocolumn,letterpaper]{article}

\usepackage{iccv}              

\definecolor{iccvblue}{rgb}{0.21,0.49,0.74}
\usepackage[pagebackref,breaklinks,colorlinks,allcolors=iccvblue]{hyperref}
\usepackage{colortbl}
\usepackage{dsfont}

\def\paperID{14295} 
\def\confName{ICCV}
\def\confYear{2025}

\title{Balancing Conservatism and Aggressiveness: \\ Prototype-Affinity Hybrid Network for Few-Shot Segmentation}

\author{
Tianyu Zou\textsuperscript{1}\ \ \ \ \ \ \ \ \ \
Shengwu Xiong\textsuperscript{2}$^{*}$ \ \ \ \ \ \ \ \ \ \
Ruilin Yao\textsuperscript{1,4,5} \ \ \ \ \ \ \ \ \ \
Yi Rong\textsuperscript{3,1}$^{*}$
\\
\textsuperscript{1} School of Computer Science and Artificial Intelligence,  Wuhan University of Technology\\
\textsuperscript{2} Interdisciplinary Artificial Intelligence Research Institute, Wuhan College\\
\textsuperscript{3} Sanya Science and Education Innovation Park, Wuhan University of Technology\\
\textsuperscript{4} School of Artificial Intelligence, University of Chinese Academy of Sciences\\
\textsuperscript{5} Foundation Model Research Center, Institute of Automation, Chinese Academy of Sciences\\
{\tt\small \{zoutianyu, xiongsw, yaoruilin, rongyi\}@whut.edu.cn}
}

\begin{document}
\maketitle
\begin{abstract}
    This paper studies the few-shot segmentation (FSS) task, which aims to segment objects belonging to unseen categories in a query image by learning a model on a small number of well-annotated support samples. Our analysis of two mainstream FSS paradigms reveals that the predictions made by prototype learning methods are usually conservative, while those of affinity learning methods tend to be more aggressive. This observation motivates us to balance the conservative and aggressive information captured by these two types of FSS frameworks so as to improve the segmentation performance. To achieve this, we propose a \textbf{P}rototype-\textbf{A}ffinity \textbf{H}ybrid \textbf{Net}work (PAHNet), which introduces a Prototype-guided Feature Enhancement (PFE) module and an Attention Score Calibration (ASC) module in each attention block of an affinity learning model (called affinity learner). These two modules utilize the predictions generated by a pre-trained prototype learning model (called prototype predictor) to enhance the foreground information in support and query image representations and suppress the mismatched foreground-background (FG-BG) relationships between them, respectively. In this way, the aggressiveness of the affinity learner can be effectively mitigated, thereby eventually increasing the segmentation accuracy of our PAHNet method. Experimental results show that PAHNet outperforms most recently proposed methods across 1-shot and 5-shot settings on both PASCAL-5$^i$ and COCO-20$^i$ datasets, suggesting its effectiveness. The code is available at: \url{https://github.com/tianyu-zou/PAHNet}

\end{abstract}    
\section{Introduction}
\label{sec:intro}


\begin{figure}[t]
    \centering
    \includegraphics[width=0.8\linewidth]{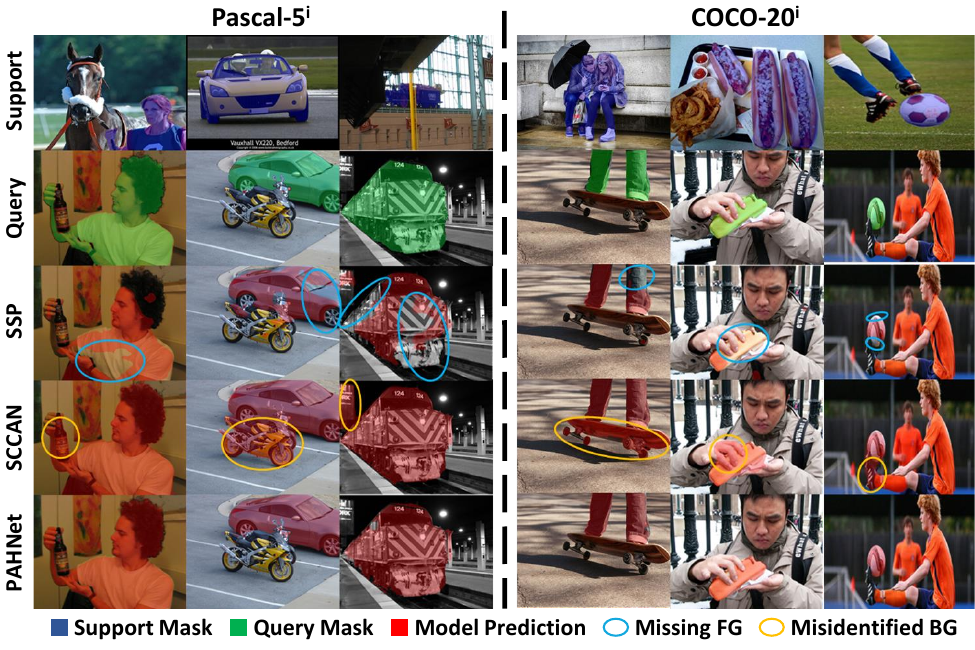}
    \caption{Prediction comparison between the prototype learning method (SSP), affinity learning method (SCCAN), and our PAHNet on PASCAL-$5^i$ and COCO-$20^i$. SSP shows significant foreground misses (\textcolor{cyan}{blue circles}), while SCCAN exhibits substantial background misactivation (\textcolor{orange}{yellow circles}). In contrast, PAHNet integrates the conservative information from the prototype learning method with the aggressive information from the affinity learning method. As a result, our PAHNet maximizes the foreground (FG) activation while reducing incorrect FG-BG matches.}
    \label{fig:motivation}
\end{figure}

Over the past decades, semantic segmentation \cite{long2015fully, lee2013pseudo, shaban2017one} has emerged as an important task in computer vision, which has attracted extensive attention from the academic community and been widely used in industrial applications. In particular, deep-learning-based semantic segmentation approaches have made rapid progress and demonstrated impressive performance. However, the success of deep learning techniques relies heavily on the availability of large-scale well-annotated data. Unfortunately, for semantic segmentation tasks, manually annotating massive samples with reliable pixel-wise labels is laborious and time-consuming. Learning deep models with such insufficient training data will inevitably lead to overfitting problems, resulting in reduced generalization ability in segmenting unseen objects. 

\begin{figure*}[t]
    \centering
    \includegraphics[width=1\linewidth]{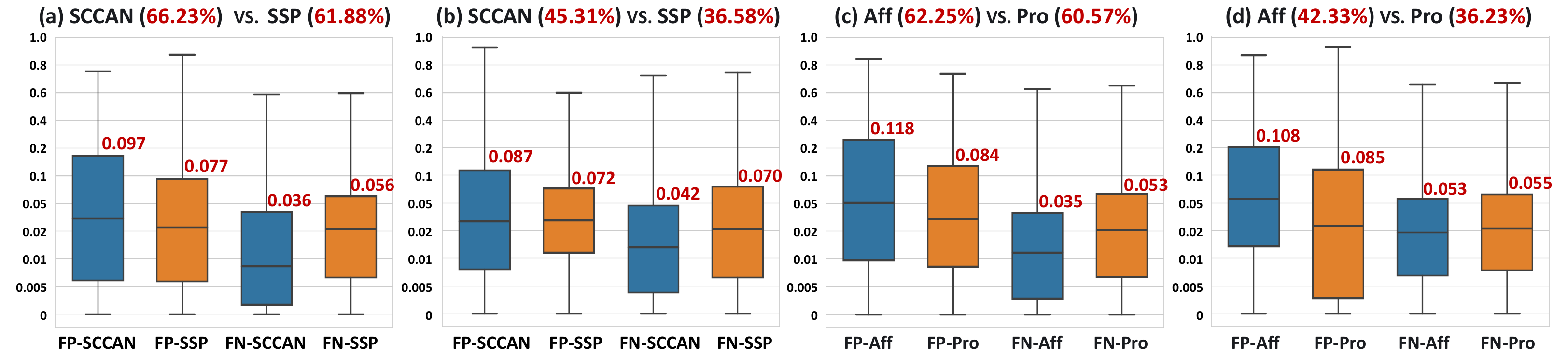}
    \caption{Comparison of average FP (False Positive) and FN (False Negative) between prototype and affinity learning methods across the four splits of PASCAL (subfigures a, c) and COCO (subfigures b, d) datasets. The red values above each box plot represent the average FP and FN percentages, and parenthetical red values in the subtigure title indicate the mIoU of these methods. Affinity learning methods exhibit higher FP but lower FN (aggressive), while prototype learning methods achieve lower FP but higher FN (conservative).}
    \label{fig:Comparison_FP_FN}
\end{figure*}

To deal with this challenge, few-shot semantic segmentation (FSS) \cite{shaban2017one, song2023comprehensive, catalano2023few} has been proposed. It emulates the ability of human intelligence that can understand new concepts by just seeing limited examples, thus enabling segmentation of unseen class images (called query samples) through utilizing useful information extracted from a small amount of annotated data (called support samples). Existing FSS methods can be roughly divided into two categories \cite{wang2023rethinking}: (1) \textbf{Prototype learning methods} \cite{wang2019panet,fan2022self,huang2023prototypical,he2023prototype,lang2022beyond} aggregate the pixel features of the target object in support images into a single or multiple representative prototypes, which are then used to guide the classification of query sample pixels; and (2) \textbf{Affinity learning methods} \cite{rakelly2018few,zhang2022mfnet,xiong2022doubly,hong2022cost,tavera2022pixel} calculate the pixel-level correlations between support and query samples (typically through the attention mechanism), and then leverage such information to enhance the features of foreground pixels in query images that are relevant with the target object.

In general, affinity learning methods tend to exhibit better performance than prototype learning ones, but we notice that the predictions produced by these two types of methods have their own distinctive characteristics. To show this, we visualize the segmentation results of SSP \cite{fan2022self} and SCCAN \cite{xu2023self} on both PASCAL-5$^i$ and COCO-20$^i$ datasets in Figure~\ref{fig:motivation}. We can find that, as a prototype learning method, SSP typically makes \textbf{conservative predictions} that miss some foreground pixels belonging to the target object (see blue circles). In contrast, the affinity learning method SCCAN can fully cover the object areas but misidentify some background pixels as foreground target (see yellow circles), thus resulting in \textbf{aggressive segmentation results}. To validate this observation, in Figure~\ref{fig:Comparison_FP_FN}(a) and (b), we display the ratio of false positive (FP) and false negative (FN) pixels predicted by SSP and SCCAN on the test data of the two datasets, respectively. As can be seen, although SCCAN significantly outperforms SSP in terms of the mIoU (\ie, mean Intersection over Union) value and the FN ratio, it consistently obtains higher average FP percentages on both datasets. This means that SSP tends to ensure the precision of its foreground predictions, while SCCAN attempts to cover more target regions. To further verify the generality of this phenomenon, we conduct the same quantitative analysis on two simplest prototype learning and affinity learning models, which are constructed by removing specific designs from SSP and SCCAN, respectively. The results are presented in Figure~\ref{fig:Comparison_FP_FN}(c) and (d), where we can again observe similar comparison situations as those between SSP and SCCAN. Therefore, we believe that \textit{the predictions made by prototype learning methods are usually conservative, whereas those of affinity learning methods are generally more aggressive.}

We analyze the reasons behind this phenomenon as follows: For prototype learning methods, their learned prototypes capture the representative (or dominant) information of entire object category, and are therefore less likely to activate background pixels in the query image that are irrelevant to the target. But due to the diversity of intra-class variants, some foreground pixels may fail to be identified by a finite number of prototypes \cite{zhang2023fgnet}, thus leading to higher FN ratios. As for affinity learning methods, the main issue lies in their pixel-level matching \cite{lu2021simpler} process. Specifically, the color or texture contexts around certain query background pixels may be similar to some support object pixels, which makes them to be incorrectly matched. As a result, the final predictions will inevitably include a portion of background areas, ultimately resulting in worse FP rates. This is called ''foreground-background (FG-BG) mismatch'' problem.

Based on the aforementioned observation and analysis, we naturally raise a question: \textit{Can we strike a balance between the conservative and aggressive predictions made by the two types of FSS frameworks so as to generate more accurate segmentation results?} This question motivates us to design and propose a novel \textbf{P}rototype-\textbf{A}ffinity \textbf{H}ybrid \textbf{Net}work (PAHNet) for FSS tasks. Our method leverages the soft predictions yielded by a pre-trained prototype learning model (called prototype predictor) as a reliable indicator for FG regions, to improve an affinity learning model (called affinity learner) by mitigating its FG-BG mismatch issue. Specifically, a Prototype-guided Feature Enhancement (PFE) module is introduced before each self-attention layer of the affinity learner to enhance the FG information of both support and query representations, under the guidance of the prototypes generated from the predictions of two models. This will strengthen the FG-FG correlations between support and query samples in the subsequent cross-attention operation. Meanwhile, PAHNet integrates an Attention Score Calibration (ASC) module within each cross-attention layer, in order to utilize the prototype predictor's predictions to re-weight the attention values and further mask the incorrect FG-BG relationships. In this way, the FG-BG mismatch problem of the affinity learner can be effectively alleviated, thus finally improving the segmentation performance of our PAHNet method (see the bottom row in Figure~\ref{fig:motivation}). Our main contributions are:

\begin{itemize}
    \item We have a new finding that the predictions of the prototype learning and the affinity learning methods have their own characteristics (one is conservative and the other is aggressive) and are generally complementary.
    \item To the best of our knowledge, our proposed PAHNet is the first unified framework that can balance the conservative and aggressive information captured by a prototype predictor and an affinity learner, thereby enhancing its segmentation performance on novel unseen classes.
    \item We design two modules (\ie, PFE and ASC), which can effectively enhance the FG information in both support and query representations, as well as suppress the mismatched FG-BG relationships between them.
    \item Extensive experiments on PASCAL-$5^i$ and COCO-$20^i$ datasets show that PAHNet outperforms most recently proposed methods, suggesting its effectiveness and superiority for addressing FSS tasks.
\end{itemize}


\begin{figure*}[t]
    \centering
    \includegraphics[width=0.85\linewidth]{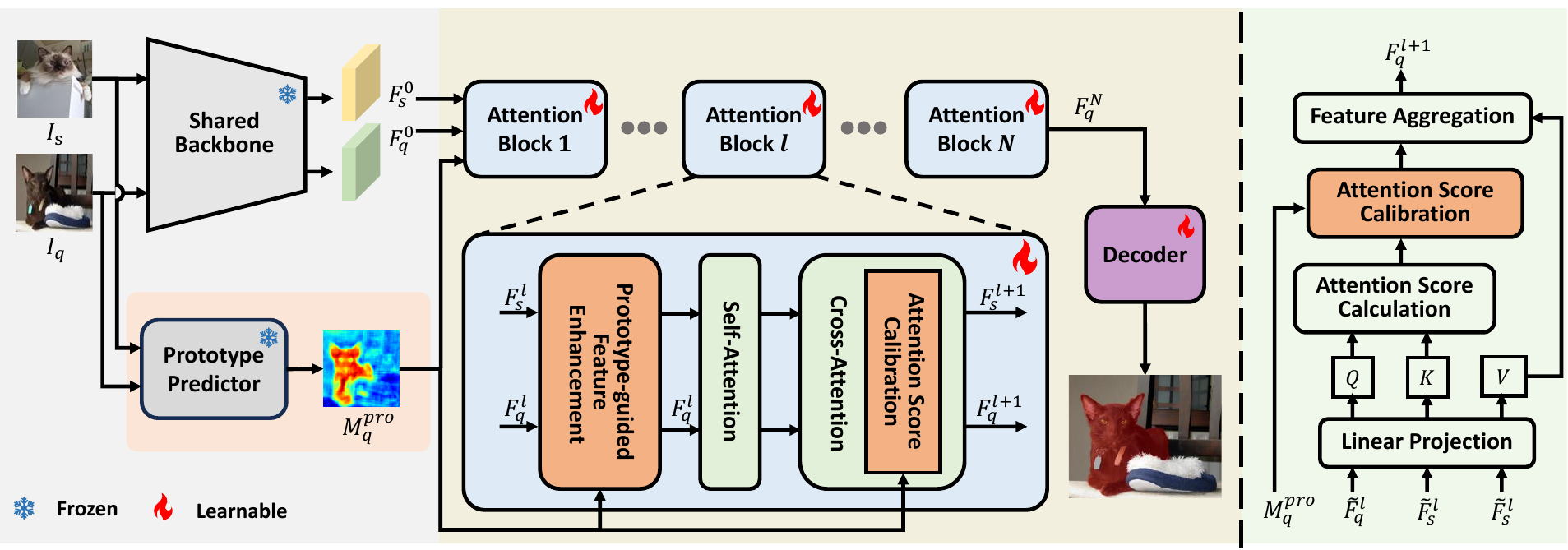}
    \caption{Architectural overview of PAHNet. It mainly consists of a pre-trained prototype predictor and a trainable affinity learner. The conservative predictions made by the prototype predictor are integrated into each attention block of the affinity learner, through a prototype-guided feature enhancement (PFE) module and a attention score calibration (ASC) module. These two modules work in synergy to enhance foreground feature discriminability and suppress foreground-background mismatches.}
    \label{fig:overall_framework}
\end{figure*}

\section{Related work}
\label{sec:related_work}
Few-Shot Segmentation (FSS) \cite{shaban2017one,wang2019panet,yang2020brinet} aims to alleviate the generalization problem on unseen classes in semantic segmentation. Existing methods primarily follow two technical paradigms, which we review in the following.

\noindent \textbf{Prototype Learning Methods.}
Prototype learning methods \cite{liu2022dynamic,hariharan2011semantic,zhang2021self} guide the segmentation of query images by extracting single or multiple class-specific prototypes from support samples. SG-One \cite{zhang2020sg} used masked average pooling to obtain a support foreground prototype and employed cosine similarity to segment the query image. While PFENet \cite{tian2020prior} introduced prior masks to coarsely localize query foreground regions and segment the query image through feature concatenation. To address the limited coverage of single prototypes, ASGNet \cite{yang2020prototype} and PMMs \cite{li2021adaptive} employed clustering to decompose targets into multi-part prototypes. Based on the intuition that pixels from the same object are more similar than those from different objects, SSP \cite{fan2022self} generated query-specific foreground (FG) and background (BG) prototypes. QPENet \cite{peng2023hierarchical} integrated query features into the generation process of foreground and background prototypes, thereby yielding customized prototypes attuned to specific queries.

\noindent {\textbf{Affinity Learning Methods.}}
To mitigate information loss and structural disruption caused by prototypes, affinity learning methods \cite{hu2019attention,zhang2022mask,wang2020few,xie2021scale} directly model support-query correlations through pixel-wise similarity learning. While direct application of pixel-wise similarity induces spurious background activation, CyCTR \cite{zhang2021few} mitigated this by developing cycle-consistent attention mechanisms to suppress feature noise in support instances. Although hypercorrelation-based methods \cite{min2021hypercorrelation,shi2022dense} employ 4D attention mechanisms to capture high-order semantic relations, they remain susceptible to foreground-background feature entanglement. SCCAN \cite{xu2023self} attributes this pathology to FG-BG mismatch in cross-attention, thus proposing self-calibration to dynamically align query FG patches with the most consistent support regions. HDMNet \cite{peng2023hierarchical} decouples the self-attention and cross-attention processes, while designing the matching module using correlation mechanisms and distillation. Recent efforts like AENet \cite{xu2025eliminating} purified FG features to improve matching accuracy, but the inherent aggressiveness of affinity propagation still risks activating erroneous BG regions.
This limitation stems from a fundamental dilemma: Aggressive affinity learning enhances FG-FG matching but amplifies FG-BG mismatches, leading to over-activation of background. Our work addresses this by hybridizing a conservative prototype predictor with aggressive affinity learning, enhancing FG features and explicitly suppressing mismatched regions while preserving accurate foreground correlations.

\section{Methodology}\label{sec:method}
The main goal of this work is to learn a segmentation model on a training set $\mathcal{D}_{\text{train}}$ with sufficient data from base classes $\mathcal{C}_{\text{base}}$, expecting that the learned model can generalize well on the FSS tasks sampled from a testing set $\mathcal{D}_{\text{test}}$ of novel classes $\mathcal{C}_{\text{novel}}$, which are not overlapped with $\mathcal{C}_{\text{base}}$, \ie, $\mathcal{C}_{\text{base}} \cap \mathcal{C}_{\text{novel}} = \emptyset$. Following the standard setup, we describe the $k$-shot segmentation task as an episode  $\mathcal{E}=\{\mathcal{S}, \mathcal{Q}\}$ that consists of a support set $\mathcal{S} = \{(I_s^{n}, M_s^{n})\}_{n=1}^k$ with $k$ samples for model adaptation and a query set $\mathcal{Q} = (I_q, M_q)$ for performance evaluation. Here, $\mathcal{S}$ and $\mathcal{Q}$ are two disjoint sets drawn from the same specific object category, satisfying $\mathcal{S} \cap \mathcal{Q} = \emptyset$. $I \in \mathbb{R}^{H \times W \times 3}$ and $M \in \{0,1\}^{H \times W}$ denote an input image and its binary segmentation mask, where $H$ and $W$ indicate their height and width, respectively. To deal with testing FFS tasks, an episodic learning strategy is usually employed. It constructs a series of episodes from $\mathcal{D}_{\text{train}}$ for model training to simulate the processing of the FFS tasks that will be encountered at the evaluation stage. In the following subsections, we will introduce our PAHNet method under the 1-shot ($k=1$) setting for clarity, yet it can be easily extended to the scenarios of $k > 1$.

\subsection{Overview of Our PAHNet}\label{sec:overview}
\vspace{-3pt}
To strike the balance between the conservative and aggressive information captured by the two types of FFS frameworks, we build our PAHNet model by combining a prototype predictor and an affinity learner, as illustrated in Figures~\ref{fig:overall_framework}. We pre-train the prototype predictor in advance (in our implementation, we directly use the officially released SSP \cite{fan2022self} model) and keep it unchanged during the training phase. As for our affinity learner, the input support and query images $\{I_s, I_q\}$ are first forwarded through a shared backbone to extract their corresponding features $\{F^{0}_{s},F^{0}_{q}\}\in \mathbb{R}^{h\times w\times d}$, where $d$ represents the channel size and $h\times w$ denotes the spatial dimensions of the features. After that, $F^{0}_{s}$ and $F^{0}_{q}$ are sequentially input into $N$ stacked attention blocks, each consisting of a self-attention layer and a cross-attention layer for capturing and exploiting the pixel-level correlations within the same image as well as between the support and query samples, respectively. Finally, the output representation $F^{N}_{q}$ of the last attention block is processed by a decoder to generate the segmentation mask $M_{q}$ of $I_q$. As analyzed in Section~\ref{sec:intro}, the prototype learning methods tend to conservatively ensure the precision of their FG predictions (with lower FP rates), so our prototype predictor can act as a reliable FG indicator for calibrating the pixel-level matching operation of the affinity learner, thus reducing FG-BG mismatches and alleviating its overly aggressive nature. To achieve this, we integrate the soft predictions ${M}_q^{pro}\in {[0,1]}^{H\times W}$ made by the prototype predictor into each attention block of our affinity learner via a Prototype-guided Feature Enhancement (PFE) module and an Attention Score Calibration (ASC) module.

\subsection{Prototype-Guided Feature Enhancement}
\begin{figure}[t]
    \centering
    \includegraphics[width=1\linewidth]{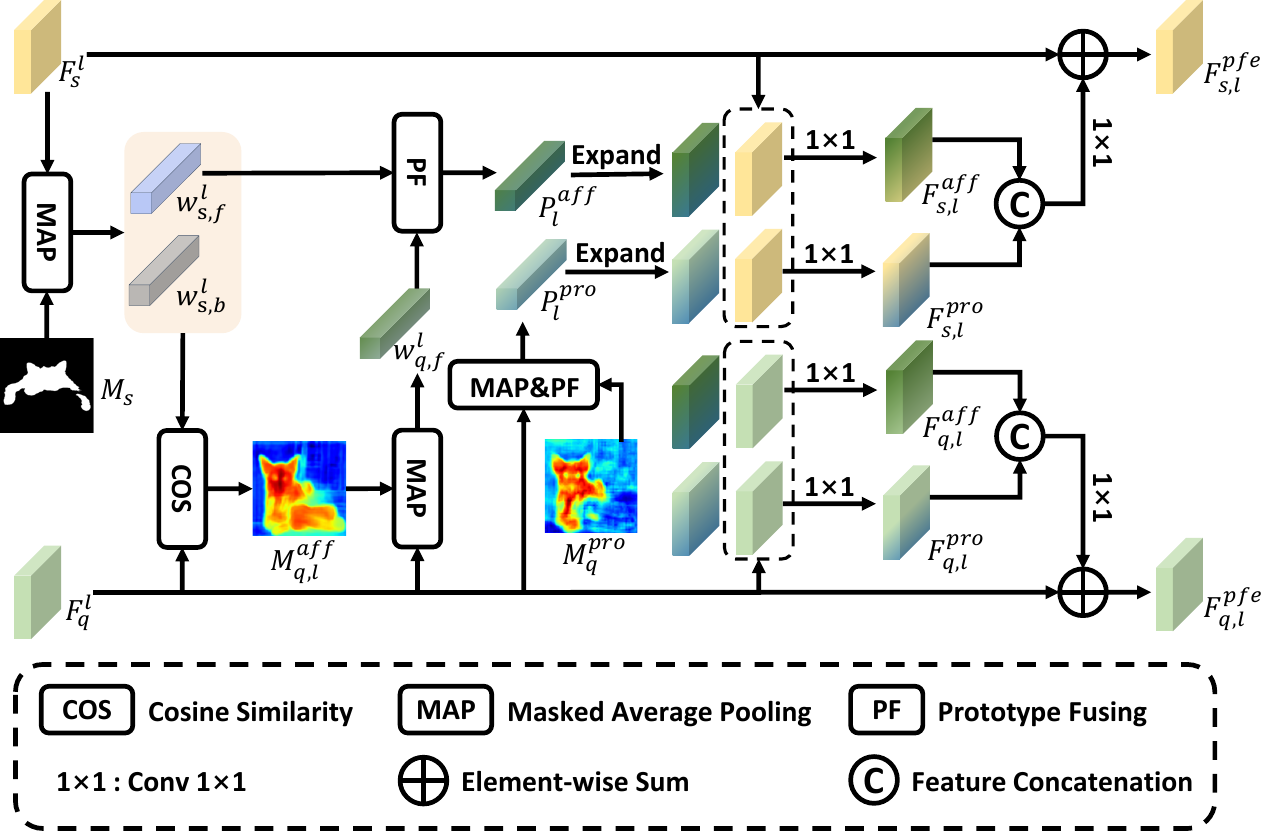}
    \caption{Illustration of our PFE module. It enhances the images features under the guidance of prototypes that are generated from the predictions of the two models.}
    \label{fig:PEF}
\end{figure}

As shown in Figure~\ref{fig:overall_framework}, the PFE module is inserted before the self-attention layer of each attention block in our affinity learner. It enhances the FG information within both support and query image representations under the guidance of a conservative prototype and an aggressive prototype, which are generated from the predictions of our prototype predictor and affinity learner, respectively. More specifically, as illustrated in Figure~\ref{fig:PEF}, for the $l$-th ($l=0,1,2,...,N-1$) attention block, we can aggregate the FG and BG information of its input support features $F^{l}_{s}$ according to the corresponding ground-truth mask $M_{s}$ as follows:
\begin{equation}\label{eq:weight}
w_{s,f}^{l}={\text{MAP}}(F^{l}_{s},M_{s}), \quad w_{s,b}^{l}={\text{MAP}}(F^{l}_{s},1-M_{s}).
\end{equation}
Here, ${\text{MAP}}(\cdot,\cdot)$ performs the masked average pooling operation on the first argument by using the mask weights given by the second argument. Based on the definition in Eq.~(\ref{eq:weight}), $w_{s,f}^{l}$ and $w_{s,b}^{l}$ can be regarded as the class weights of a binary linear classifier for distinguishing the FG and BG pixels in $F^{l}_{s}$. Therefore, we can apply them to the query image features $F^{l}_{q}$ to produce the following predictions:
\begin{equation}\label{aff_mask}
    \!\!M_{q,l}^{aff}(i,j)= \frac{\exp({\text{Cos}}(F^{l}_{q}(i,j),w_{s,f}^{l})/\tau)}{\sum_{k\in\{f,b\}}\exp({\text{Cos}}(F^{l}_{q}(i,j),w_{s,k}^{l})/\tau)},  \\
\end{equation}
where ${\text{Cos}}(\cdot,\cdot)$ calculates the cosine similarity between the two inputs and $F^{l}_{q}(i,j)$ represents the feature vector of the $(i,j)$-th pixel in $F^{l}_{q}$. $\tau>0$ is a temperature parameter that controls the scale of probability logits. With $M_{q,l}^{aff}$ defined above, the FG information of $F^{l}_{q}$ can also be summarized as $w_{q,f}^{l}={\text{MAP}}(F^{l}_{q},M_{q,l}^{aff})$. Then, by fusing $w_{s,f}^{l}$ and $w_{q,f}^{l}$, we obtain a prototype which is able to indicate the FG areas in both support and query features:
\begin{equation}\label{eq:prototype}
    p_l^{aff}=\alpha\cdot w_{s,f}^{l}+(1-\alpha)\cdot w_{q,f}^{l}.
\end{equation}
Here, $\alpha =(\text{Cos}(w_{s,f}^{l},w_{q,f}^{l})+1)/2$ is an adaptive parameter that balances the relative importance of $w_{s,f}^{l}$ and $w_{q,f}^{l}$. This means that if $w_{q,f}^{l}$ is more different from $w_{s,f}^{l}$, it will contain more specific FG information that is not captured by the support features, and therefore it should be assigned a larger $\alpha$ to incorporate such information into $p_l^{aff}$. Next, we utilize this prototype to enhance $F^{l}_{s}$ and $F^{l}_{q}$ as follows:
\begin{equation}\label{eq:enhancement1}
    F_{k,l}^{aff}=\text{Conv}_{1\times1}([F^{l}_{k};\:p_l^{aff}]), \:\: k\in\{s,q\},
\end{equation}
where $\text{Conv}_{1\times1}(\cdot)$ denotes the convolution operation with a kernel size of ${1\times1}$, and $[\cdot;\:\cdot]$ indicates the feature concatenation along the channel dimension. However, due to the predictive aggressiveness of the affine learner, $M_{q,l}^{aff}$ may be perturbed by some BG noises, which further affects the effectiveness of $F_{s,l}^{aff}$ and $F_{q,l}^{aff}$. To this end, we perform the same operations on the predictions ${M}_q^{pro}$ of our prototype predictor as those applied on $M_{q,l}^{aff}$, resulting in another conservative version of enhanced features $F_{s,l}^{pro}$ and $F_{q,l}^{pro}$. Through the following fusion operation, we get the final enhanced support and query representations:  
\begin{equation}\label{eq:enhancement2}
    {F}^{pfe}_{k,l}=\text{Conv}_{1\times1}([F_{k,l}^{aff};\:F_{k,l}^{pro}])+F^l_{k}, \:\: k\in\{s,q\}.
\end{equation}
Afterwards, these PFE-enhanced features are separately fed into the multi-head self-attention layer, which generates the inputs $\tilde{F}^l_{s}$ and $\tilde{F}^l_{q}$ for the subsequent cross-attention layer.

\subsection{Attention Score Calibration}

\begin{figure}
    \centering
    \includegraphics[width=0.9\linewidth]{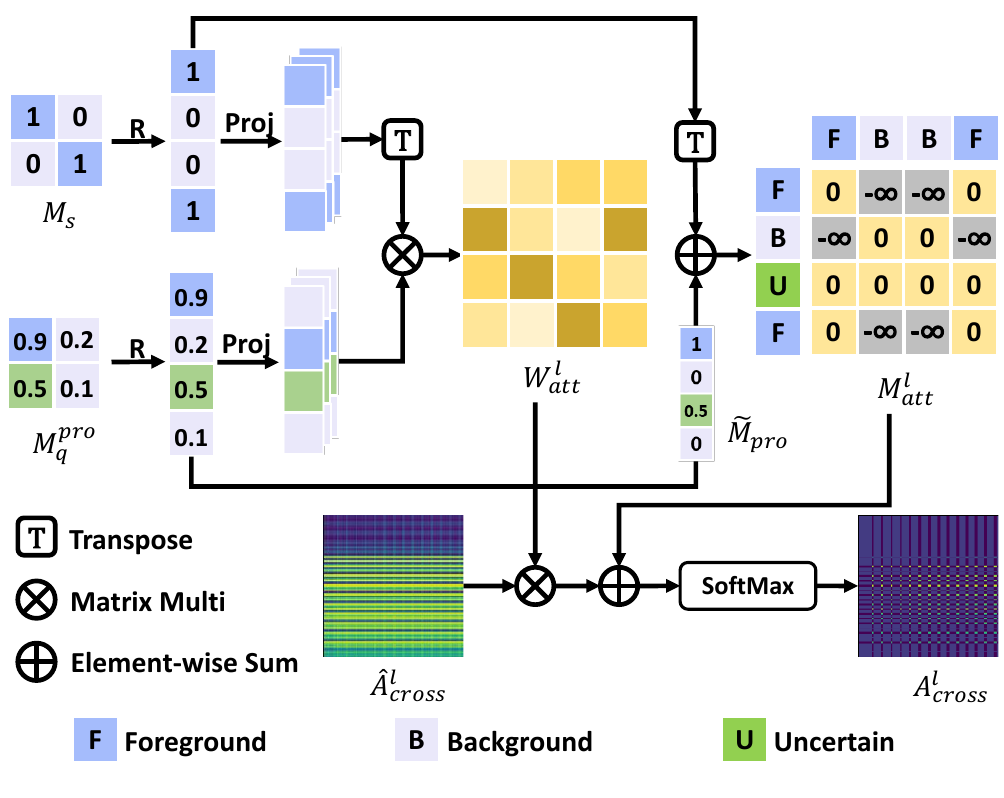}
    \caption{Illustration of our ASC module. It re-weights the cross-attention scores and then masks the erroneous FG-BG mismatches.}
    \label{fig:ASC}
\end{figure}

While the PFE module enhances the FG information in $\tilde{F}^l_{s}$ and $\tilde{F}^l_{q}$, and thus strengthens their within- and cross-image FG-FG relationships, it does not explicitly reduce the FG-BG mismatches between them, such incorrect correlations can still degrade the final segmentation performance. To solve this issue, we integrate an ASC module (as shown in Figure~\ref{fig:ASC}) into the cross-attention layer of each attention block. It utilizes $M_s$ and ${M}_q^{pro}$ to re-weight the attention scores (\ie, similarities between Query and Key embeddings) to lower the impact of potential FG-BG correlations. And for those certainly mismatched FG-BG relationships, ASC directly filters out them by further masking their corresponding attention scores. Specifically, to do this, ASC first constructs a re-weight matrix as follows: 
\begin{equation}\label{eq:reweight}
    W^l_{att}= [\text{Proj}^l(\varphi({M}_q^{pro}))]\cdot[\text{Proj}^l(\varphi(M_s))]^{\text{T}},
\end{equation}
where $\varphi: \mathbb{R}^{h \times w} \mapsto \mathbb{R}^{hw \times 1}$ is a spatially flatten operation, and $\text{Proj}: \mathbb{R}^{hw\times 1} \mapsto \mathbb{R}^{hw \times d_c}$ denotes a linear projector that maps the input into a $d_c$-dimensional space. According to Eq.~(\ref{eq:reweight}), $W^l_{att} \in \mathbb{R}^{hw \times hw}$ calculates the pixel-level correlations between the conservative predictions ${M}_q^{pro}$ for the query image and the groud-truth mask $M_s$ of the support image. Through the projection function $\text{Proj}(\cdot)$, the predictive and mask values indicating the FG regions (\ie, larger values in ${M}_q^{pro}$ and \textbf{1}s in $M_s$) and the BG regions (\ie, smaller values in ${M}_q^{pro}$ and \textbf{0}s in $M_s$) are projected into different vector clusters. Consequently, $W^l_{att}$ will assign smaller weights to the potential FG-BG mismatches between $\tilde{F}^l_{s}$ and $\tilde{F}^l_{q}$, so as to suppress their negative effects.

\begin{table*}[t]
    \centering
    \resizebox{0.81\textwidth}{!}{
    \begin{tabular}{l|c|cccccc|cccccc}
        \toprule
        & & \multicolumn{6}{c|}{1-shot} & \multicolumn{6}{c}{5-shot} \\
        Methods & Venue & $5^0$ & $5^1$ & $5^2$ & $5^3$ & Mean & FB-IoU & $5^0$ & $5^1$ & $5^2$ & $5^3$ & Mean & FB-IoU  \\
        \midrule
        FECANet \cite{liu2023fecanet} & TMM'23 & 69.2 & 72.3 & 62.4 & 65.7 & 67.4 & 78.7 & 72.9 & 74.0 & 65.2 & 67.8 & 70.0 & 80.7  \\ 
        TEM \cite{chen2024transformer} & IJCAI’24 & 67.9 & 74.3 & 61.1 & 64.6 & 67.0 & - & 72.0 & 76.4 & 64.5 & 69.1 & 70.5 & -  \\ 
        QPENet \cite{cong2024query} & TMM'24 & 65.2 & 71.9 & 64.1 & 59.5 & 65.2 & 76.7 & 68.4 & 74.0 & 67.4 & 65.2 & 68.8 & 80.0 \\
        RiFeNet \cite{bao2024relevant} & AAAI’24& 68.4 & 73.5 & 67.1 & 59.4 & 67.1 & - & 70.0 & 74.7 & 69.4 & 64.2 & 69.6 & -  \\ 
        AENet \cite{xu2025eliminating} & ECCV’24& 71.3 & 75.9 & 68.6 & 65.4 & 70.3 & 81.2 & 73.9 & 77.8 & 73.3 & 72.0 & 74.2 & 84.5  \\ 
        PMNet \cite{chen2024pixel} & ECCV’24& 67.3 & 72.0 & 62.4 & 59.9 & 65.4 & -  & 73.6 & 74.6 & 69.9 & 67.2 & 71.3 & - \\
        ABCBNet \cite{zhu2024addressing} &CVPR’24& 72.9 & 76.0 & 69.5 & 64.0 & 70.6 & - & 74.4 & 78.0 & 73.9 & 68.3 & 73.6 & - \\
        HMNet \cite{xu2024hybrid} &NIPS’24& 72.2 & 75.4 & 70.0 & 63.9 & 70.4 & 81.6 & 74.2 & 77.3 & 74.1 & 70.9 & 74.1 & 84.4  \\ 
        \rowcolor{blue!12}
        SCCAN \cite{xu2023self} &ICCV’23& 68.3 & 72.5 & 66.8 & 59.8 & 66.8 & 77.7 & 72.3 & 74.1 & 69.1 & 65.6 & 70.3 & 81.8  \\  
        \rowcolor{red!12}
        SCCAN+PAHNet & Ours & \textbf{73.9} & 74.5 & \textbf{73.4} & 64.5 & \textbf{71.6} & \textbf{82.6} & 75.7 & 75.8 & \textbf{78.3} & 71.4 & 75.3 & 85.2  \\ 
        \rowcolor{blue!12}
        HDMNet \cite{peng2023hierarchical} &CVPR’23& 71.0 & 75.4 & 68.9 & 62.1 & 69.4 & - & 71.3 & 76.2 & 71.3 & 68.5 & 71.8 & -  \\ 
        \rowcolor{red!12}
        HDMNet+PAHNet & Ours& 72.1 & \textbf{76.4} & 71.0 & \textbf{66.6} & 71.5 & 82.4 & \textbf{75.8} & \textbf{78.6} & 76.9 & \textbf{73.2} & \textbf{76.1} & \textbf{85.7}  \\ 
        \bottomrule
    \end{tabular}
    }
    \caption{Performance comparisons on PASCAL-$5^i$ in terms of mIoU and FB-IoU.  ``$5^i$'' shows the mIoU scores of 5 novel classes in fold $i$, ``Mean'' denotes the averaged mIoU score across all four folds. The best results are highlighted in bold.}
    \label{tab:pascal}
\end{table*}

\begin{table*}
    \centering
    \resizebox{0.81\textwidth}{!}{
    \begin{tabular}{l|c|cccccc|cccccc}
    \toprule
        & &\multicolumn{6}{c|}{1-shot} & \multicolumn{6}{c}{5-shot} \\
        Methods &Venue& $20^0$ & $20^1$ & $20^2$ & $20^3$ & Mean & FB-IoU & $20^0$ & $20^1$ & $20^2$ & $20^3$ & Mean & FB-IoU  \\
        \hline
        FECANet \cite{liu2023fecanet} & TMM'23& 38.5 & 44.6 & 42.6 & 40.7 & 41.6 & 69.6 & 44.6 & 51.5 & 48.4 & 45.8 & 47.6 & 71.1  \\ 
        TEM \cite{chen2024transformer} &IJCAI’24& 44.2 & 51.5 & 47.8 & 46.5 & 47.5 & - & 49.3 & 58.6 & 56.9 & 53.8 & 54.7 & -  \\ 
        QPENet \cite{cong2024query} &TMM'24& 41.5 & 47.3 & 40.9 & 39.4 & 42.3 & 67.4 & 47.3 & 52.4 & 44.3 & 44.9 & 47.2 & 69.5 \\
        RiFeNet \cite{bao2024relevant} &AAAI’24& 39.1 & 47.2 & 44.6 & 45.4 & 44.1 & - & 44.3 & 52.4 & 49.3 & 48.4 & 48.6 & -  \\ 
        AENet \cite{xu2025eliminating} &ECCV’24& 45.4 & 57.1 & 52.6 & 50.0 & 51.3 & 74.4 & 52.7 & 62.6 & 56.8 & 56.1 & 57.1 & \textbf{78.5}  \\ 
        PMNet \cite{chen2024pixel} &ECCV’24& 39.8 & 41.0 & 40.1 & 40.7 & 40.4 & - & 50.1 & 51.0 & 50.4 & 49.6 & 50.3 & - \\
        ABCBNet \cite{zhu2024addressing} &CVPR’24& 44.2 & 54.0 & 52.1 & 49.8 & 50.0 & - & 50.5 & 59.1 & 57.0 & 53.6 & 55.1 & - \\
        HMNet \cite{xu2024hybrid} &NIPS’24& 45.5 & \textbf{58.7} & 52.9 & \textbf{51.4} & 52.1 & 74.5 & 53.4 & \textbf{64.6} & \textbf{60.8} & 56.8 & 58.9 & 77.6  \\ 
        \rowcolor{blue!12}
        SCCAN \cite{xu2023self} &ICCV’23& 40.4 & 49.7 & 49.6 & 45.6 & 46.3 & 69.9 & 47.2 & 57.2 & 59.2 & 52.1 & 53.9 & 74.2  \\ 
        \rowcolor{red!12}
        SCCAN+PAHNet &Ours& 43.8 & 56.8 & 51.6 & 48.6 & 50.2 & 73.7 & 51.1 & 63.2 & 59.6 & 55.9 & 57.5 & 76.8  \\ 
        \rowcolor{blue!12}
        HDMNet \cite{peng2023hierarchical} &CVPR’23& 44.8 & 54.9 & 50.0 & 48.7 & 49.6 & 72.1 & 50.9 & 60.2 & 55.0 & 55.3 & 55.3 & 74.9  \\ 
        \rowcolor{red!12}
        HDMNet+PAHNet &Ours& \textbf{45.2} & 58.2 & \textbf{54.4} & 51.2 & \textbf{52.3} & \textbf{75.2} & \textbf{53.0} & 64.1 & \textbf{60.8} & \textbf{58.7} & \textbf{59.2} & 78.1  \\ 
        \bottomrule
    \end{tabular}
    }
    \caption{Performance comparisons on COCO-$20^i$ in terms of mIoU and FB-IoU. ``$20^i$" shows the mIoU scores of 5 novel classes in fold $i$, ``Mean" denotes the averaged mIoU score across all four folds. The best results are highlighted in bold.}
    \label{tab:coco}
\end{table*}

In addition, by setting two thresholding parameters, we can categorize the predicted FG probabilities in ${M}_q^{pro}$ into three different groups as follows:
\begin{equation} \label{eq:indicator}
\tilde{M}_q^{pro}(i,j)= 
    \begin{cases}
    1, & \text{if } {M}_q^{pro}(i,j) \ge \gamma_{fg} \\
    0, & \text{if }{M}_q^{pro}(i,j) \leq \gamma_{bg} \\
    {M}_q^{pro}(i,j), & \text{otherwise}
    \end{cases}
\end{equation}
With this definition, $\gamma_{fg}>0$ and $\gamma_{bg}>0$ recognize the FG and BG areas with high predictive confidence, respectively. Therefore, we can utilize $M_s$ and $\tilde{M}_q^{pro}$ to identify and filter the FG-BG correlations that are definitely mismatched between the support and query representations:
\begin{equation} \label{eq:mask_score}
\!\!M^l_{att}(i,j)= 
    \begin{cases}
    -\infty,\!\!\!\!& \text{if } \varphi(\tilde{M}_q^{pro})(i)+\varphi(M_s)(j)=1 \\
    0,& \text{otherwise}
    \end{cases}
\end{equation}
Here, $\varphi(M)(i)$ is the $i$-th element of the vector $\varphi(M)$. The first condition in Eq.~(\ref{eq:mask_score}) is satisfied only if $\varphi(\tilde{M}_q^{pro})(i)=0$ and $\varphi(M_s)(j)=1$  or vice versa, therefore can be used to indicate the highly determined FG-BG mismatches. Finally, with $W^l_{att}$ and $M^l_{att}$, the ASC module calibrates the cross-attention scores of the $l$-th block as follows:
\begin{equation}\label{attention_weight}
A^l_{cross} = \text{Softmax}(W_{att} \odot \frac{Q^l (K^l)^{\text{T}}}{\sqrt{d}} + M^l_{att}),
\end{equation}
where $\odot$ denotes the Hadamard product (\ie, element-wise multiplication). $\text{Softmax}(\cdot)$ represents the softmax function along the column dimension. $Q^l$ and $K^l$ are Query and Key embeddings that are derived from $\tilde{F}^l_{q}$ and $\tilde{F}^l_{s}$, respectively. As can be seen, if $M^l_{att}(i,j)=-\infty$, then $A^l_{cross}(i,j)=0$, which means that the $i$-th Query token will not absorb information from the $j$-th Key token, thus removing the FG-BG mismatch between them. After that, the output representations of the $l$-th attention block can be formulated as:
\begin{equation}\label{attention_weight}
 F^{l+1}_{s}=\tilde{F}^l_{s}, \quad F^{l+1}_{q}= A^l_{cross}V^l,
\end{equation}
where $V^l$ is also projected from $\tilde{F}^l_{s}$. Note that the cross-attention layer only changes the features of the query sample, while $\tilde{F}^l_{s}$ of the support image is directly output to the next block. To train our PAHNet model, we adopt the same loss function as in AENet \cite{xu2025eliminating}. \textbf{Please refer to the Supplementary Material for more details of our training loss.}

\begin{figure*}[t]
    \centering
    \includegraphics[width=0.85\linewidth]{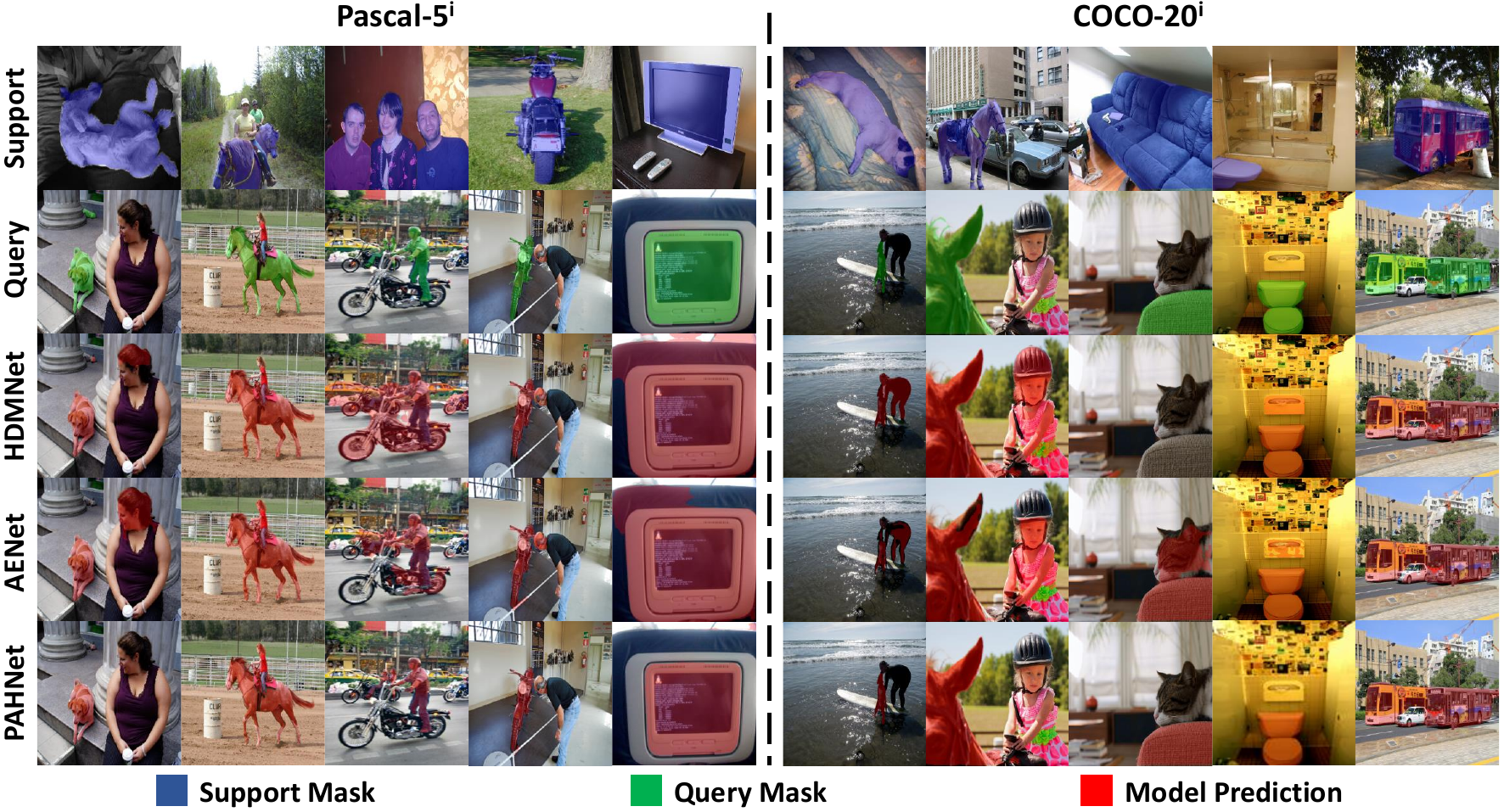}
    \caption{Qualitative comparison under 1-shot setting on PASCAL-$5^i$ and COCO-$20^i$, showing results of HDMNet and its improved variants AENet and our proposed PAHNet.}
    \label{fig:qualitative_results}
\end{figure*}

\section{Experiments}
\subsection{Experimental Settings}
\noindent \textbf{Datasets.} We evaluate our PAHNet method on two widely used datasets, including PASCAL-$5^i$ \cite{shaban2017one} and COCO-$20^i$ \cite{nguyen2019feature}. \textbf{PASCAL-$5^i$} is created from PASCAL VOC 2012 \cite{everingham2010pascal} with additional annotations from SDS \cite{hariharan2011semantic}, while \textbf{COCO-$20^i$} is built from MSCOCO \cite{lin2014microsoft}. The object categories in both datasets are evenly divided into four folds, and the experiments are conducted in a cross-validation manner. For each fold, we randomly sample 1,000 and 4,000 episodes from PASCAL-$5^i$ and COCO-$20^i$ for testing. 

\noindent {\textbf{Evaluation metrics.}} Following the standard FSS protocols \cite{wang2024rethinking,fan2022self,xu2023self}, two evaluation metrics are adopted: (1) \textbf{mIoU} computes the average IoU over all foreground classes, reflecting overall segmentation accuracy; (2) \textbf{FB-IoU} separately measures foreground and background IoUs, then averages them to assess the class balance performance. 

\noindent {\textbf{Implementation Details.}} 
Since our PAHNet method does not change the core attention operations in the affinity learner, it can be flexibly integrated into different affinity learning methods. To validate this compatibility, we combine PAHNet with two recently proposed methods, namely SCCAN \cite{xu2023self} and HDMNet \cite{peng2023hierarchical}. And as mentioned in Section~\ref{sec:overview}, we directly utilize the officially released SSP \cite{fan2022self} model as 
our prototype learner. Same as previous works \cite{lang2022learning,fan2022self}, the ResNet-50 \cite{he2016deep} pretrained on ImageNet \cite{deng2009imagenet} is adopted as the shared backbone for image feature extraction. For a fair comparison, we use the same decoder architecture as SCCAN and HDMNet, and strictly follow their training configurations (including the settings of data augmentation, optimizer, batch size, learning rate, and \etc), when working with each of them. As for the hyperparameters, we set the temperature parameter in Eq.~(\ref{aff_mask}) as $\tau=0.1$, the thresholding parameters in Eq.~(\ref{eq:indicator}) as $\gamma_{fg}=0.7$ and $\gamma_{bg}=0.3$. All the experiments are conducted on a single NVIDIA A100 GPU with 40 GB memory. 

\subsection{Comparison with State-of-the-Art Methods}
\noindent \textbf{Quantitative Comparison.} 
To validate the effectiveness of our PAHNet, we compare its segmentation performance under both the 1-shot and 5-shot settings with those obtained by other recent methods, including AENet \cite{xu2025eliminating}, HMNet \cite{xu2024hybrid} and \etc. The segmentation results for the PASCAL-$5^i$ and COCO-$20^i$ datasets are presented in Table~\ref{tab:pascal} and Table~\ref{tab:coco}, from which we can get the following observations: \textbf{1) By equipping our PAHNet, the performance of both SCCAN and HDMNet is significantly improved}. Specifically, for the PASCAL-$5^i$ dataset, PAHNet improves SCCAN by 4.8\% and 5.0\%, HDMNet by 2.1\% and 4.3\% under the two settings, respectively. Whereas for the COCO-$20^i$ dataset, SCCAN+PAHNet and HDMNet+PAHNet outperforms their corresponding baselines by 3.9\%, 2.7\% and 3.6\%, 3.9\% for the 1-shot and 5-shot settings. These results demonstrate that it is reasonable and effective to mitigate the predictive aggressiveness of the affinity learner through the conservative predictions from a prototype predictor. \textbf{2) The performance gains achieved by PAHNet are typically more pronounced in the 5-shot setting than in the 1-shot setting}. For example, the 5-shot gains over the HDMNet baseline are 4.3\% (PASCAL-$5^i$) and 3.9\% (COCO-$20^i$), exceeding the 1-shot improvements by 2.2\% and 1.2\%, respectively. It is mainly because richer support samples will allow our prototype predictor to produce more accurate predictions. This will provide better guidance for our PFE and ASC modules to more appropriately enhance the FG information and reduce the FG-BG mismatches. \textbf{3) On both datasets, PAHNet outperforms most recently proposed methods across the two settings.} On COCO-$20^i$, the mean IoUs achieved by HDMNet+PAHNet are slightly better than HMNet but surpass other competing methods by a large margin of at least 1.0\% (vs. AENet) and 2.1\% (vs. AENet). Such improvements become more significant on the PASCAL-$5^i$ dataset, outperforming HMNet by 1.1\% and 2.0\%, as well as exceeding other methods by more than 0.9\% (vs. ABCBNet \cite{zhu2024addressing}) and 1.9\% (vs. AENet). All the above results suggest the superiority of our PAHNet method in solving FSS tasks.

\noindent \textbf{Qualitative Comparison.} To better understand the effectiveness of our method, we also visualize the segmentation results of HDMNet, AENet, and our HDMNet+PAHNet in Figure~\ref{fig:qualitative_results} to make a qualitative comparison.
As can be seen, since AENet improves HDMNet by amplifying the FG ratio in feature representations, it can recover the ``sofa'' in the third column of COCO-$20^i$ that HDMNet completely fails to identify. We can also find that results achieved by our PAHNet method are more accurate than those of HDMNet and AENet, where only PAHNet correctly recognizes and clearly filters the BG pixels of the ``cat'' in the same image. Additionally, the similar results can also be observed when segmenting other query samples. Such superior performance of PAHNet may be attributed to the ASC module that decouples the mismatched FG-BG correlations from the legitimate object contexts, thus enabling the discriminative enhancement while suppressing the BG interference.

\begin{table}[t]
    \centering
    \resizebox{1.0\linewidth}{!}{
    \begin{tabular}{l|rrrrr}
    \toprule
        Method & $5^0$ & $5^1$ & $5^2$ & $5^3$ & Mean  \\ \hline
        SSP (Prototype Learning) & 60.3 & 65.8 & 67.0 & 52.0 & 61.2  \\
        SCCAN (Affinity Learning)  & 68.3 & 72.5 & 66.8 & 59.8 & 66.8  \\
        HDMNet (Baseline) & 71.0& 75.4& 68.9& 62.1& 69.4\\
        PAHNet (HDMNet+SSP) & 72.1& 76.4& 71.0& 66.6& 71.5\\
        PAHNet (HDMNet+SCCAN) & 71.9& 75.7 & 69.4 & 64.7 & 70.4  \\
        \bottomrule
    \end{tabular}
    }
    \caption{Ablation study on the effect of prototype predictor.}
    \label{tab:Auxiliary Branches}
\end{table}

\begin{table}[t]
    \centering
    \resizebox{0.85\linewidth}{!}{
    \begin{tabular}{ccccccc}
    \toprule
        PFE & ASC & $5^0$ & $5^1$ & $5^2$ & $5^3$ & Mean  \\ \hline
        ~ & ~ & 69.6 & 73.2 & 69.3 & 61.6 & 68.4  \\ 
        \textbf{\checkmark} & ~ &  72.3 & 73.7 & 71.8 & 63.3 & 70.3 \\
        ~ & \textbf{\checkmark} & 73.1 & 74.1 & 73.1 & 63.6 & 71.0 \\
        \textbf{\checkmark} & \textbf{\checkmark} & 73.9 & 74.5 & 73.4 & 64.5 & 71.6  \\
        \bottomrule
    \end{tabular}
    }
    \caption{Ablation study on the effect of PFE and ASC modules.}
    \label{tab:Component-wise}
\end{table}

\subsection{Ablation Study}
We conduct a series of ablation studies to investigate the impact of different components in our method on segmentation performance. Note that the experiments in this section are conducted with the combination of SCCAN and BAM as the baseline on the PASCAL-$5^i$ dataset under the 1-shot setting unless specified otherwise.

\noindent \textbf{Effect of Prototype Predictor.}
To investigate how the conservative predictions made by our prototype predictor influence the effectiveness of PAHNet, we take HDMNet as the baseline and replace the original prototype predictor (\ie, SSP) in our implementation with an affinity learning model (\ie, SCCAN). It can be observed from Table~\ref{tab:Auxiliary Branches} that, even though SCCAN performs much better than SSP (66.8\% vs. 61.2\%), its  performance improvements over the baseline is 1.1\% lower than that of SSP (70.4\% and 71.5\%), when combining with HDMNet through our method. This demonstrates that the performance gains of PAHNet stem not merely from additional predictor model, but also from the balance of prediction conservatism and aggressiveness.

\noindent \textbf{Effect of PFE and ASC modules.}
We study the effect of our PFE and ASC modules by removing each of them from our PAHNet. As shown in Table \ref{tab:Component-wise}, the baseline mIoU starts at 68.4\% and increases to 70.3\% and 71.0\% with the integration of the PFE and ASC modules, respectively. When both PFE and ASC modules are used together, the mIoU further improves to 71.6\% (+3.2\%), demonstrating the effectiveness of both modules in improving FFS results.

\begin{figure}[t]
    \centering
    \includegraphics[width=0.9\linewidth]{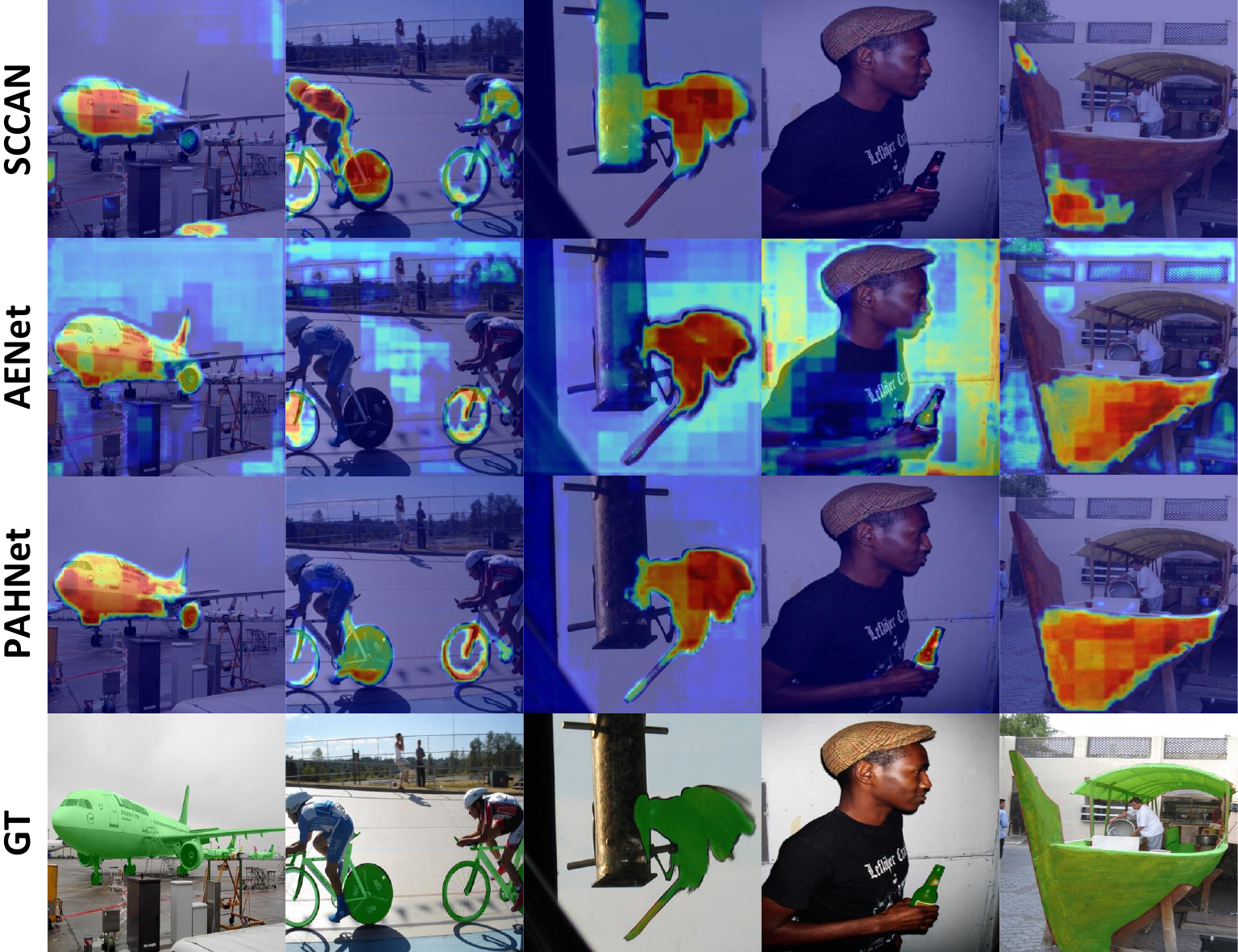}
    \caption{Grad-CAM visualization of the output features from the last attention block of different methods.}
    \label{fig:grad}
\end{figure}

We also adopt the Grad-CAM technique to generate the visualization results of the baseline SCCAN, AENet, and our SCCAN+PAHNet in Figure~\ref{fig:grad}. It can be seen that the output features of SCCAN and AENet often incorrectly focus on the BG areas, which may lead to segmentation errors. In contrast, after enhanced by our PFE and ASC modules, the image features of PAHNet pay more attentions on the target object and are thus more discriminative. \textbf{Please refer to the Supplementary Material for more experimental results and additioanl discussions.}

\section{Conclusion}
In this work, based on the observation that the prototype learning methods tend to make conservative predictions and those of affinity learning methods are usually aggressive. We propose a novel hybrid framework called PAHNet to achieve a balance between the predictive conservatism and aggressiveness of a prototype projector and an affinity learner. By introducing the PFE and ASC modules in each attention block, PAHNet effectively enhances the FG information in image features and mitigates the FG-BG mismatches. Extensive experiments on PASCAL-$5^i$ and COCO-$20^i$ demonstrate the effectiveness of PAHNet in addressing FFS tasks under various settings.

\newpage
\section*{Acknowledgements}
This work was partially supported by the Young Scientists Fund of the National Natural Science Foundation of China (Grant No. 62306219), the National Key Research and Development Program of China (Grant No. 2022ZD0160604), the National Natural Science Foundation of China (Grant No. 62176194), and the Key Research and Development Program of Hubei Province (Grant No. 2023BAB083), and was also supported in part by computing resources from the Wuhan Supercomputing Center and the Wuhan Artificial Intelligence Computing Center.

\bibliographystyle{ieeenat_fullname}
\bibliography{main}

\end{document}